# Feature Selection Using Classifier in High Dimensional Data

Shivani Pathak and Vijendra Singh

Department of Computer Science, Faculty of Engineering and Technology,
Mody Institute of Technology and Science, Lakshmangarh,
Rajasthan, India-332311

**Abstract**
Feature selection is frequently used as a pre-processing step to machine learning. It is a process of choosing a subset of original features so that the feature space is optimally reduced according to a certain evaluation criterion. The central objective of this paper is to reduce the dimension of the data by finding a small set of important features which can give good classification performance. We have applied filter and wrapper approach with different classifiers QDA and LDA respectively. A widely-used filter method is used for bioinformatics data i.e. a univariate criterion separately on each feature, assuming that there is no interaction between features and then applied Sequential Feature Selection method. Experimental results show that filter approach gives better performance in respect of Misclassification Error Rate.

## 1. INTRODUCTION

*1.1 Feature Selection*

In modeling problems, the analyst is often faced with more predictor variables than can be usefully employed. Consider that the size of the input space (the space defined by the input variables in a modeling problem) grows exponentially. Cutting up each input variable's scale into, say, 10 segments implies that a single-input model will require 10 examples for model construction, while a model with two input variables will require 100 (= 10 x 10), and so forth (assuming only one training example per cell). Assuming that the inputs are completely uncorrelated, six input variables, by this criterion, would require 1 million examples. In real problems, input variables are usually somewhat correlated, reducing the number of needed examples, but this problem still explodes rapidly, and these estimates can be considered somewhat conservative in that perhaps more than one example should be available from each cell. Given this issue, data miners are often faced with the task of selecting which predictor variables to keep in the model. This process goes by several names, the most common of which are subset selection, attribute selection and feature selection. Many solutions have been proposed for this task, though none of them are perfect, except on very small problems. Most such solutions attack this problem directly, by experimenting with predictors to be kept. When applied in biology domain, the technique is also called discriminative gene selection, which detects influential genes based on DNA microarray experiments. By removing most irrelevant and redundant features from the data, feature selection helps improve the performance of learning models by:

- Alleviating the effect of the curse of dimensionality.
- Enhancing generalization capability.
- Speeding up learning process.
- Improving model interpretability.

*1.2. Methods used for feature selection*

- Filter Method

- Wrapper Method
- Hybrid Method

Filter Method: The filter approach requires the statistical analysis of the feature set only for solving the FS task without utilizing any learning model [6]. The filters work fast using a simple measurement, but its result is not always satisfactory.

Wrapper Method: The wrapper approach involves with the predetermined learning model, selects features on measuring the learning performance of the particular learning model.

Hybrid Method: The hybrid approach attempts to take advantage of the filter and wrapper approaches [6, 12]. It is often found that, hybrid technique is capable of locating a good solution, while a single technique often traps into an immature solution. In the Hybrid approach filter models can be chosen as the preliminary screening to remove the most redundant or irrelevant features for ex. F-score and information gain. These two resulted feature sets are combined together as the pre-processed feature set for fine tuning. This step is called the combination model [7]. Finally, the wrapper model is applied to improve the classification accuracy, and this is the fine-tuning step.

## 2. RELATED WORK

Stream wise feature selection [1] considers new features sequentially and adds them to a Predictive model. Here all features are known in advance. Thus, the flexibility that stream wise regression provides to dynamically decide which features to generate and add to the feature stream provides potentially large savings in computation. Empirical tests show that for the smaller UCI data sets where stepwise regression can be done, stream wise regression gives comparable results to stepwise regression or techniques such as decision trees, neural networks, or SVMs.

A new hybrid genetic algorithm for feature selection is introduced called as HGAFS [2]. This algorithm selects salient feature subset within a reduced size. HGAFS incorporates a new local search operation that is devised and embedded in HGA to fine-tune the search in feature selection process. The aim is to guide the search process so that the newly generated offspring can be adjusted by the less correlated (distinct) features consisting of general and special characteristics of a given dataset. Thus, this algorithm reduces redundancy of information among the selected features.

A stochastic algorithm based on the GRASP [3] meta-heuristic is been proposed, with the main goal of speeding up the feature subset selection process, basically by reducing the number of wrapper evaluations to carry out. GRASP is a multi-start constructive method which constructs a solution in its first stage, and then runs an improving stage over that solution. Several instances of the proposed GRASP method are experimentally tested and compared with state-of-the-art algorithms over 12 high-dimensional datasets. The complexity, in terms of wrapper evaluations carried out, of the proposed GRASP algorithm comes from the following parameters: n, the number of predictive variables; m the cardinality of the subset selected for the constructive step; k the number of iterations carried out by the grasp algorithm; and, the wrapper FSS method selected for the improving step.

A hybrid algorithm, SAGA [4], is proposed for this task. SAGA combines the ability to avoid being trapped in a local minimum of simulated annealing with the very high rate of convergence of the crossover operator of genetic algorithms, the strong local search ability of greedy algorithms and the high computational efficiency of generalized regression neural networks. The performance of SAGA and well-known algorithms on synthetic and real datasets are compared.

### 3. FEATURE SELECTION ALGORITHM WITH CLASSIFIER

Reducing the number of features (dimensionality) is important in statistical learning. For many data sets with a large number of features and a limited number of observations, such as bioinformatics data, usually many features are not useful for producing a desired learning result and the limited observations may lead the learning algorithm to overfit to the noise. This paper shows how to perform sequential feature selection, which is one of the most popular feature selection algorithms. It also shows how to use holdout and cross-validation to evaluate the performance of the selected features.

*3.1 Discriminant Analysis Method (Classifier)*

- QDA(Quadratic Discriminant Analysis)
- LDA(Linear Discriminant Analysis)

Quadratic discriminant analysis (QDA) is closely related to linear discriminant analysis (LDA), where it is assumed that there are only two classes of points and that the measurements are normally distributed. Unlike LDA however, in QDA there is no assumption that the covariance of each of the classes is identical. When the assumption is true, the best possible test for the hypothesis that a given measurement is from a given class is the likelihood ratio test.

Linear discriminant analysis (LDA) is used in statistics, pattern recognition and machine learning to find a linear combination of features which characterizes or separates two or more classes of objects or events. The resulting combination may be used as a linear classifier or, more commonly, for dimensionality reduction before later classification. LDA explicitly attempts to model the difference between the classes of data.

*3.2 Validations Used*

*3.2.1 Hold-out Validation*

The holdout method is the simplest kind of cross validation. The data set is separated into two sets, called the training set and the testing set. The function approximator fits a function using the training set only. Then the function approximator is asked to predict the output values for the data in the testing set (it has never seen these output values before). The errors it makes are accumulated as before to give the mean absolute test set error, which is used to evaluate the model.

*3.2.2 Cross Validation*

Cross-validation [1], sometimes called rotation estimation, is a technique for assessing how the results of a statistical analysis will generalize to an independent data set. In $k$-fold cross-validation, the original sample is randomly partitioned into $k$ subsamples. Of the $k$ subsamples, a single subsample is retained as the validation data for testing the model, and the remaining $k - 1$ subsamples are used as training data. The cross-validation process is then repeated $k$ times (the *folds*), with each of the $k$ subsamples used exactly once as the validation data. The $k$ results from the folds then can be averaged (or otherwise combined) to produce a single estimation.

In stratified *k*-fold cross-validation, the folds are selected so that the mean response value is approximately equal in all the folds. In the case of a dichotomous classification, this means that each fold contains roughly the same proportions of the two types of class labels.

*3.4 Algorithm (Sequential Feature Selection)*

Sequential forward selection (SFS) and sequential backward selection (SBS) are commonly used algorithms, usually in combination with a wrapper evaluator. Both algorithms have O ($n^2$) worst-case complexity, but especially in high-dimensional datasets only SFS is used because the evaluations are simpler (fewer variables in the selected subset).

SFS is a bottom up search procedure that adds new features to a feature set one at a time until the final feature set is reached. Suppose we have a set of $d_1$ features $X_{d1}$. For each of the feature yet not selected $\xi_j$ (i.e. in X- $X_{d1}$) the criterion function $J_j = J(X_{d1} + \xi_j)$ is evaluated. The feature that yields the maximum value of $J_j$ is chosen as the one that is added to the set $X_{d1}$ thus at each stage the variable is chosen that, when added to the current set, maximises the selection criterion. The feature set is initialised to the null set. When the best improvement makes the feature set worst, or when the maximum allowable number of feature is reached, the algorithm terminates. Here J can be given by

$J = X_k^T . S_k^{-1} . X_k$ where $X_k$ is a k dimensional vector and $S_k$ is a k*k positive definite matrix when k features are used. Following algorithm explains the whole procedure, at each stage of the search, sets of subsets are generated for evaluation using the cross validation procedure. Variable $\xi_j$ is chosen for which $J(X- \xi_j)$ is the largest. The new set is (X- $\xi_j$). This process is repeated until the set of required cardinality remains.

Cross Validation algorithm for selection of the best set of features in a sequential search procedure is explained as follows:

**ALGORITHM**

    **1.** Divide the data into training and test set.

    **2.** Specify the search strategy (here SFS).

    **At each stage of the algorithm:**

(a) Generate subsets of feature for evaluation.

(b) Cross-validation procedure:

    i. Split the training data into e.g. 10 equal parts, ensuring that all classes are represented in each part; use nine parts for training and remaining one part for testing.

    ii. Train the classifier model for each nested subset of variable, $S_h$, on each subset, k, of the training data in turn, testing on the remaining part. Obtain the performance for e.g. error rate, CV(h, k), h=1, …,p; k=1, … , 10

    iii. Average the result

$$CV(h) = \frac{1}{10}\sum_{k} CV(h,k); \text{CV: Cross Validation}$$

    **3.** Select the smallest feature subset, $S_{h*}$, such that CV (h) is optimal or near optimal.

    **4.** Evaluate on test dataset using the feature subset $S_{h*}$, training on the entire training set and evaluating performance on the test set.

## 4. EXPERIMENTAL RESULTS

We used a clinical high dimensional data here which have 216 rows and 4000 features. The data variable consists of 216 observations with 4000 features. Here, we divide data into a training set of size 160 and a test set of size of size 56. We use Quadratic Discriminant Analysis (QDA) and Linear Discriminant Analysis (LDA) separately on the same dataset as the classification algorithm. In this paper we explained one example of a filter method and one example of a wrapper method. A widely-used filter method for bioinformatics data is to apply a univariate criterion separately on each feature, assuming that there is no interaction between features. For example, we might apply the t-test on each feature and compare p-value (or the absolute values of t-statistics) for each feature as a measure of how effective it is at separating groups. In order to get a general idea of how well-separated the two groups are by each feature, we plot the empirical cumulative distribution function (CDF) of the p-values. In the graph obtained after classification(Fig 1 (a);(b) ) there are about 35% of features having p-values close to zero and 50% of features having p-values smaller than 0.05, meaning there about 2000 features among the original 4000 features that have strong discrimination power.

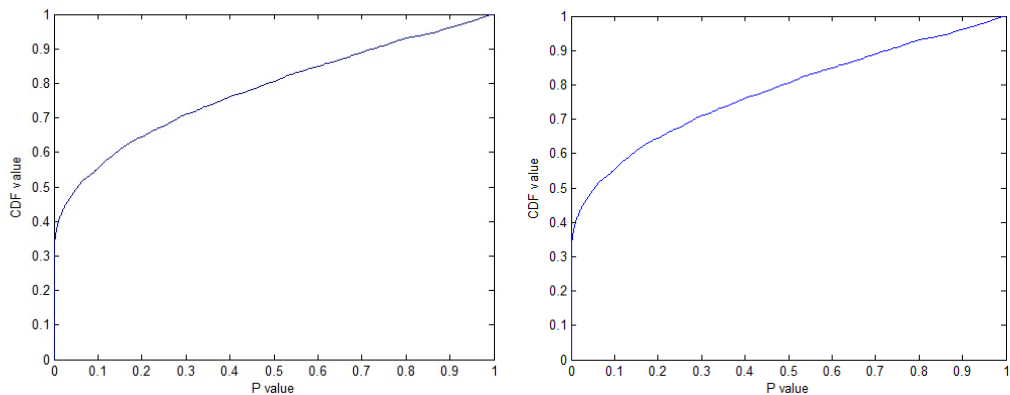

Fig. 1. (a) QDA                                    (b) LDA

We can sort these features according to their p-values (or the absolute values of the t-statistic) and select some features from the sorted list. However, it is usually difficult to decide how many features are needed unless one has some domain knowledge or the maximum number of features that can be considered has been dictated in advance based on outside constraints. One quick way to decide the number of needed features is to plot the MCE (misclassification error, i.e., the number of misclassified observations divided by the number of observations) on the test set as a function of the number of features. Since there are only 160 observations in the training set, the largest number of features for applying QDA and LDA is limited, otherwise, there may not be enough samples in each group to estimate a covariance matrix. Actually, for the data used in this paper, the holdout partition and the sizes of two groups dictate that the largest allowable number of features for applying QDA is about 70. Now we compute MCE for various numbers of features between 5 and 70 and show the plot of MCE as a function of the number of features. In order to reasonably estimate the performance of the selected model, it is important to use the 160 training samples to fit the QDA and LDA model and compute the MCE on the 56 test observations. This simple filter feature selection method gets the smallest MCE on the test set when 20 features and 60 features are used in QDA and LDA respectively. The plot (Fig. 2 (a), (b)) shows overfitting occurs when more than these features are used.

The training set is used to select the features and to fit the QDA model, and the test set is used to evaluate the performance of the finally selected feature. During the feature selection procedure, to evaluate and to compare the performance of the each candidate feature subset, we apply stratified 10-fold cross-validation to the training set. Then we use the filter results as a pre-processing step to select features. For example, we select 150 features here. We apply forward sequential feature selection on these 150 features. It stops when the first local minimum of the cross-validation MCE is found. The algorithm may have stopped prematurely. Sometimes a smaller MCE is achievable by looking for the minimum of the cross-validation MCE over a reasonable range of number of features. The cross-validation MCE (QDA), as shown in fig.3(a) reaches the minimum value when 9 features are used and this curve stays flat over the range from 9 features to 27 features (except for a slight increase when 22 features are used). Also the curve goes up when more than 28 features are used, which means overfitting occurs there. It is usually preferable to have fewer features, so here we pick 9 features, and calculated the final Misclassification Error that comes out to be 0.0714. Same procedure when applied using LDA Misclassification Error comes out to be 0.0893 when 9 features are picked as shown in fig.3 (b).

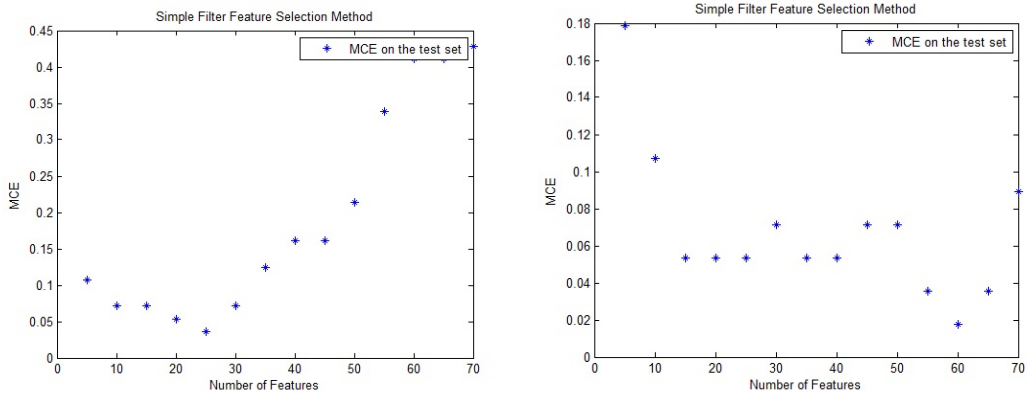

Fig. 2. (a) QDA  (b) LDA

Table 1. Misclassification Error

| Approach | QDA | LDA |
| --- | --- | --- |
| Simple Feature Selection | 0.0536 | 0.0200 |
| Sequential Feature Selection | 0.0714 | 0.0893 |

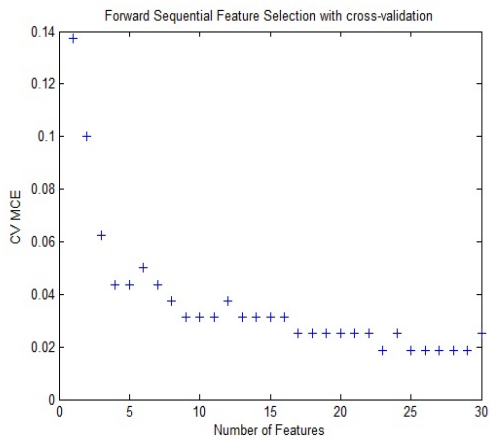 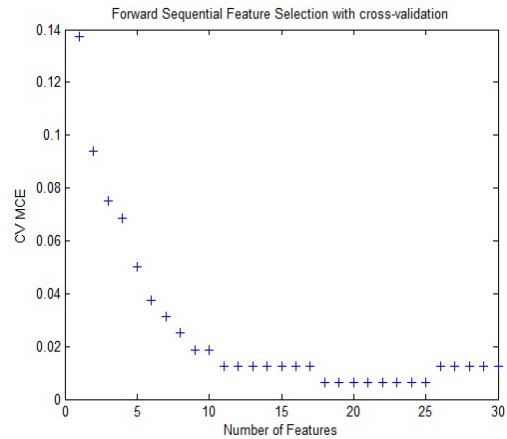

Fig. 3. (a) QDA  (b) LDA

## 5. CONCLUSION

Misclassification error rate (the number of misclassified observations divided by the number of observations) is calculated in this paper to decide the number of needed features using holdout validation and cross-validation. In order to reasonably estimate the performance of the selected model, we used 160 training samples to fit the model and compute the MCE on the 56 test observations.

Simple Filter Feature Selection method gets the smallest MCE on the test set when 20 features are used in QDA and 15 features are used in LDA. The plot shows overfitting occurs when more than these features are used.